\documentclass{article}

\usepackage[preprint]{neurips_2023}

\usepackage[utf8]{inputenc} 
\usepackage[T1]{fontenc}    
\usepackage{url}            
\usepackage{booktabs}       
\usepackage{amsfonts}       
\usepackage{nicefrac}       
\usepackage{microtype}      
\usepackage{xcolor}         

\usepackage{amsmath,amssymb,mathrsfs,subfigure,graphicx,marvosym,algorithm,algorithmic}
\usepackage{footmisc}

\usepackage{tikz}

\usetikzlibrary{positioning}
\setcitestyle{square,numbers}
\title{Prompt Engineering Through the Lens of Optimal Control}
\author{Yifan Luo,\footnotemark[1]\ \ \footnotemark[4]
\And Yiming Tang,\footnotemark[1]\ \ \footnotemark[4]
\And Chengfeng Shen,\footnotemark[1]\ \ \footnotemark[4]
\And Zhennan Zhou,\footnotemark[2]
\And Bin Dong\footnotemark[3]
}

\begin{document}
\bibliographystyle{unsrt}
\maketitle

\footnotetext[1]{School of Mathematical Science, Peking University.\\ (luoyf@pku.edu.cn; 2201110049@pku.edu.cn; 1900010740@pku.edu.cn)}
\footnotetext[2]{Beijing International Center for Mathematical Research, Peking University. (zhennan@bicmr.pku.edu.cn)}
\footnotetext[3]{Beijing International Center for Mathematical Research, Peking University; Center for Machine Learning Research, Peking University; National Biomedical Imaging Center, Peking University. \\(dongbin@math.pku.edu.cn)}
\footnotetext[4]{Equal contribution.}

\begin{abstract}
    Prompt Engineering (PE) has emerged as a critical technique for guiding Large Language Models (LLMs) in solving intricate tasks. Its importance is highlighted by its potential to significantly enhance the efficiency and effectiveness of human-machine interaction. As tasks grow increasingly complex, recent advanced PE methods have extended beyond the limitations of single-round interactions to embrace multi-round interactions, which allows for a deeper and more nuanced engagement with LLMs. In this paper, we propose an optimal control framework tailored for multi-round interactions with LLMs. This framework provides a unified mathematical structure that not only systematizes the existing PE methods but also sets the stage for rigorous analytical improvements. Furthermore, we extend this framework to include PE via ensemble methods and multi-agent collaboration, thereby enlarging the scope of applicability. By adopting an optimal control perspective, we offer fresh insights into existing PE methods and highlight theoretical challenges that warrant future research. Besides, our work lays a foundation for the development of more effective and interpretable PE methods.
\end{abstract}

\section{Introduction}

Prompt engineering (PE) first emerged in the field of Large Language Models (LLMs) in 2020, as researchers realized that well-designed prompts could significantly enhance the capabilities of LLMs without additional model training \cite{brown2020language,antoun2020aragpt2,logan2021cutting, muktadir2023brief}. The development of PE can be contextualized within the larger scope of natural language programming \cite{mihalcea2006nlp, miller1981natural}-- an increasingly prevalent paradigm that allows for the manipulation of computational systems through natural language, thus offering a more intuitive alternative to traditional programming languages.
Much like the transition from machine language to higher-level languages like C marked a significant leap in expressive power and ease of use, prompt engineering -- or natural language programming in a broader sense -- represents a further evolutionary leap, making it easier than ever to instruct machines in performing complex tasks.
When implemented properly, PE can yield dramatic performance improvements, particularly in the context of advanced LLMs such as GPT-4 and Claude. In these sophisticated models, the gap between well-engineered and poorly conceived prompts can be stark, reinforcing the critical role of effective PE in leveraging the full potential of LLMs.

Initially, the focus of PE was on single-round prompting, a mechanism suited for relatively straightforward tasks. However, as the need for more complex problem-solving through natural language programming became evident, the field saw a shift towards more intricate forms of engagement, such as multi-round and even multi-agent interactions with LLMs \cite{zhou2023leasttomost,Zhou2022LargeLM,yao2023tree,besta2023graph}. This evolution in PE bears a striking resemblance to the historical trajectory of optimal control theory \cite{kirk2004optimal}, which itself originated from the need for point-to-point trajectory optimization and later expanded its scope to accommodate dynamic systems with feedback mechanisms.

The growing complexity of multi-round PE interactions presents significant challenges. Traditional PE approaches \cite{brown2020language,Wei2022ChainOT,kojima2023large} often rely on heuristic or empirical methods that, while effective in specific scenarios, lack a systematic foundation amenable to rigorous analysis. This highlights the pressing need for a unified mathematical framework that can serve as a descriptive foundation and facilitate optimization of multi-round PE dynamics.

The primary aim of this paper is to introduce a novel optimal control framework tailored for multi-round interactions with LLMs. Unlike previous works with limited theoretical scopes \cite{liu2023pre,dong2023survey,qiao2023reasoning,bhargava2023whats}, our approach offers a comprehensive mathematical structure for the systematic design, analysis, and optimization of PE methods, broadening its applicability to include ensemble and multi-agent strategies.

Adopting an optimal control perspective holds the promise of evolving PE along a trajectory similar to that of optimal control theory itself. Initial methodologies in PE mainly focused on single-round prompts, comparable to point-to-point trajectory optimization problems \cite{zhang2022automatic, Gao2022PALPL, Shi2023PromptSO}. As optimal control theory incorporated feedback mechanisms for handling complex systems, our framework is designed to accommodate both single-round and multi-round interactions. This shift aims to offer a coherent understanding of the dynamics governing these intricate exchanges and to foster innovative applications transcending current limitations.

To realize these objectives, our methodology employs optimal control to conceptualize multi-round LLM interactions. While acknowledging existing gaps in mathematical rigor due to poorly understood metrics in discrete language spaces, the framework aims to serve as a unified lens for qualitatively evaluating existing PE techniques. Thus, it lays the groundwork for potential improvements in PE by providing an intuitive, structurally coherent approach to model extended dialogic interactions.

Contributions of this paper are summarized as follows:
\begin{enumerate}
\item We introduce a novel optimal control formulation that unifies a wide range of existing methods under a single mathematical framework. This provides a rigorous foundation for analyzing and improving prompt design.
\item We highlight theoretical challenges revealed by the framework, specifically regarding the formalization and optimization of multi-round interactions. While complex, these issues offer exciting directions for future studies to deepen the mathematical understanding of PE.
\item Our perspective yields valuable insights into the inherent capabilities and limitations of current techniques. These could catalyze innovations in PE, pushing the boundaries of human-computer interaction.
\item We extend the framework to ensemble PE methods and multi-agent PE, serving as an important stepping stone for studying complex interactions with LLMs.
\end{enumerate}
We note that the primary aim of this paper is not to present new theoretical results or algorithmic improvements substantiated by experiments; rather, we introduce an optimal control framework to systematize and interpret existing PE methods, thereby laying the groundwork for future rigorous analysis in the domain of PE.

The remainder of this paper is structured as follows. Section 2 elaborates on the pivotal concepts in PE and introduces the optimal control framework designed to systematize PE. In this section, we also shed light on the significance of multi-round interactions, highlighting the challenges and opportunities that multi-round PE presents. In Section 3, we review several well-established PE methods, integrating them into the proposed optimal control framework and elucidating the new insights that emerge from this integration. Section 4 is dedicated to extended PE methodologies, such as ensemble and multi-agent PE strategies. We illustrate how minor adaptations to the proposed framework can accommodate these more sophisticated, yet potent, PE methods. The paper concludes with Section 5, where we summarize our contributions.

\section{A General Framework for Prompt Engineering}
\label{sec:framework}

\subsection{Concepts and Terminologies}
In the context of LLMs, a \textbf{prompt} serves as a starting point for an interaction with the LLM. It could be a question, a statement, or a command that are given in natural language. The quality of prompts are important because they have a strong effect on the quality of the LLM's responses. Well-designed prompts can lead to useful and accurate responses, while a poorly designed ones may result in irrelevant or wrong responses.

In this paper, our main focus is on \textbf{multi-round interactions} with LLMs. In these situations, a user interacts with the model multiple times to complete one specific task. For multi-round interactions, the user gives a series of prompts and later prompts can be influenced by LLM's previous responses. This approach allows the user to get more detailed information and helps the model to handle tasks that are too complex for a single interaction.

\textbf{Prompt engineering (PE)} is the process of designing and refining a sequence of prompts to be used in a multi-round interaction with an LLM, with the goal of eliciting a satisfactory final response. By "satisfactory", we mean that the final response should score high on some measurable standard. In Section \ref{sec:sub:general_framework}, we will show how the idea of finding refined prompts can be described as an optimal control problem.

\subsection{Why is Multi-Round Interaction Necessary?}
It is important to emphasize the significance of engaging in multiple rounds of interaction for obtaining high-quality responses from LLMs \cite{wang2023mint}. One could argue that a single well-crafted prompt should be sufficient if the LLM is highly advanced. However, this is often not achievable. Even with a sophisticated LLM, crafting the ideal prompt can be challenging, especially when the user is not an expert in the subject matter at hand.

For instance, consider a medical diagnosis scenario where a user is experiencing symptoms and seeks advice from an LLM. An initial prompt might capture a basic description of the symptoms; however, lacking medical expertise, the user may miss pertinent information. For example,  the user might overlook some crucial connections between their symptoms and lifestyle habits or other factors. These unstated contextual factors could be vital for the LLM to provide an accurate or useful response in a single interaction round. 

\cite[Chapter 4]{lee2023ai} provides a detailed example of GPT-4 assisting a physician with a neonatal diagnosis. Initially, the physician outlines a set of symptoms, eliciting from GPT-4 a list of four possible conditions. In subsequent rounds of dialogue, supplemented with extra clinical details including ultrasound images and hormone level data, GPT-4 refines its assessment and identify the most probable diagnosis among the four possibilities. This collaborative process culminates in the accurate detection of a remarkably rare disorder, with an incidence of less than one in every 100,000 newborns. Given the technical nature and length of the original conversation, it has been excluded from this text. Readers seeking an in-depth understanding are directed to consult \cite[Chapter 4]{lee2023ai}.

In a multi-round interaction, both the user and the LLM have the opportunity for a more extended exchange of information. The user can adapt their prompts based on the LLM's previous responses, adding details or context that were initially lacking. Similarly, the LLM might ask clarifying questions that help guide the user to provide additional, more relevant information.

Through this iterative process of exchanging information, the specialized knowledge of the LLM assists the user in crafting more effective prompts, enabling a more productive dialogue. In summary, engaging in multi-round interactions enhances the cooperation between users and LLMs. The back-and-forth conversations facilitate more contextual and nuanced exchanges that are vital to unlocking the full potential of LLMs.

\subsection{General Framework}
\label{sec:sub:general_framework}

Let us define the text space as $\mathcal{Z}$. Elements in $\mathcal{Z}$ are compositions of some tokens selected from the token vocabulary $\mathcal{T}$, i.e. $z=[t_1, \ldots, t_m]\in \mathcal{Z}$ where $t_k\in\mathcal{T}$. Later in this paper, we also use $[z_1z_2]$ to represent the text obtained by concatenating $z_1$ and $z_2$ one after another.

Under these notations, a given LLM can be mathematically modeled as a transformation over \( \mathcal{Z} \):
\begin{equation*}
    \operatorname{LLM}: \begin{aligned}
        \mathcal{Z} &\to\mathcal{Z}\\
        z^{p} &\mapsto z^{r}
    \end{aligned}
\end{equation*}
where $z^p$ stands for prompt and $z^r$ stands for response.

\textbf{Remark.} While our current formulation treats LLMs as deterministic transformations for the sake of conceptual clarity, it is noteworthy that all LLMs operate in a stochastic setting. The inherent variability due to stochastic sampling in text generation can yield different outputs for the same input prompt. Interestingly, certain PE methods such as ensemble methods can benefit from this stochastic property. We will explore this with more details in section \ref{sec:extensions}. Nevertheless, the primary insights and conclusions drawn in this paper remain valid in both deterministic and stochastic settings.

Given a task description (or a query) \( z^q \), our objective is to obtain an optimal response, where optimality is measured using an evaluation function \( f \):
\begin{equation*}
    f: \begin{aligned}
        \mathcal{Z} &\to\mathbb{R}\\
        z &\mapsto f(z;z^q)
    \end{aligned}
\end{equation*}

The purpose of PE is to find a sequence of prompts \( \{z_t^p\}_{t=1}^{\tau} \) that lead to an optimal response. Formally, this can be viewed as an optimal control problem:
\begin{equation}
    \begin{aligned}
        \max_{\tau} & \max_{z_t^{p}\in\mathcal{P}_t} f(z_\tau^{r};z^q)+R(\tau)\\
        s.t.\ & z_t^{r}=\operatorname{LLM}(z_t^{p})\\
    \end{aligned}
    \label{def:PE_basic}
\end{equation}
where $R(\tau)$ is a regularization term. An example of $R(t)$ is the following function, which enforces a maximum interaction limit of $T$:
$$
R(t)=\begin{cases}
    0,\quad & t\leq T\\
    -\infty, & t>T
\end{cases}
$$

In this formulation, \( \{\mathcal{P}_t\}_{t=1}^{\tau} \) is a sequence of \textbf{prompt candidate sets} that expand over rounds: 
\begin{equation*}
     \mathcal{P}_t \subset \mathcal{P}_{t+1}.
\end{equation*}
This expansion encapsulates the user's growing understanding through iterative interactions. Initially, the prompt candidate set contains only general queries. However, as the conversation progresses, as the user get access to $\{z^{r}_s\}_{s=1}^t$, the set \( \mathcal{P}_{t+1} \) enlarges because of the additional information acquired through the conversations, i.e. $\mathcal{P}_{t+1} =\mathcal{P}_{t}\cup\{\text{new prompts based on $z^r_t$}\}$. Thus, it may include more specific and relevant prompts, culminating in a sufficiently large set that holds an optimal prompts for task completion. 

Some readers may question the necessity of constraining the prompt candidate set $\mathcal{P}_t$ rather than setting it equal to the entire prompt space $\mathcal{Z}$ for all $t$. It is imperative to understand that the cardinality of $\mathcal{Z}$ is overwhelmingly large, making optimization within this comprehensive space computationally infeasible. In our optimal control framework, the enlargement of the prompt candidate set $\mathcal{P}_t$ is permitted only after the associated information for new prompts has undergone scrutiny. This ensures that the size of $\mathcal{P}_t$ remains within computationally manageable bounds. Importantly, this feature of having an enlarging candidate set is a novel aspect that departs from traditional optimal control theory. Although it introduces additional analytical and computational complexities, such a formulation is naturally motivated by the practical requirements of PE as well as other real-world scenarios. A thorough discussion on this matter is deferred to Section 3.1.

\begin{figure}[h]
    \centering
    \includegraphics[width=14cm]{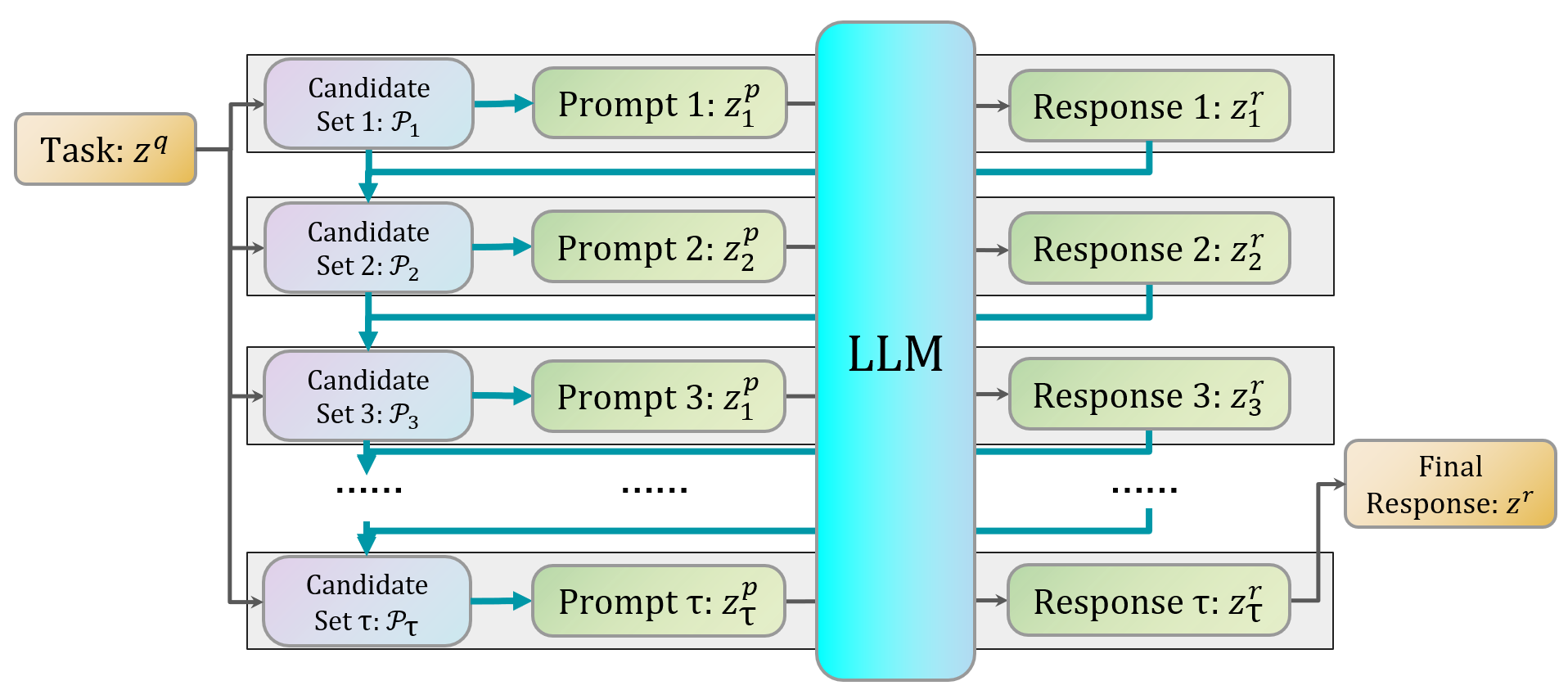}
    \caption{
    The general framework of multi-round PE. In our optimal control formulation \eqref{def:PE_basic}, the task (or query) is denoted by $z^q$, the prompt candidate sets are denoted as $\mathcal{P}_t$, which is updated based on preceding response $z_{t-1}^r$. The answer is the final response from the LLM. We use bold arrows to demonstrate two procedures: choosing $z_t^p$ from $\mathcal{P}_t$ and the enlarging of $\mathcal{P}_t$. We can interpret $\mathcal{P}_t$ as the embodiment of our "action space" when prompting.
    }
    \label{fig0}	
\end{figure}

To summarize, within our framework, PE is essentially the formulation and resolution of problem (\ref{def:PE_basic}). More specifically, it encompasses the following tasks:
\begin{itemize}
    \item Determining a suitable evaluation function \( f \);
    \item Establishing an update rule for the prompt candidate set \( \mathcal{P}_t \);
    \item Solving the resultant optimal control problem, i.e. choosing $z_t^p$ from $\mathcal{P}_t$.
\end{itemize}
Here, the first two tasks pertain to problem formulation, while the final task focuses on solving the underlying problem. Figure \ref{fig0} shows a schematic diagram of our framework.

\subsection{Potentials and Challenges in Prompt Engineering}

Within our framework, the multi-round interaction with an LLM constitutes a dynamic system, and PE is framed as a control problem defined over the dynamic system. From this optimal control perspective, we will discuss the potentials that PE may achieve and the challenges that PE presents. 

\subsubsection*{Potentials}

Let us first briefly discuss the characteristics of LLM which help gain insight into PE methodologies via the optimal control framework. In our view, the key traits of LLMs that enable effective PE are their immense knowledge capacity and inherent variability. 

With knowledge capacity, an adept LLM contains extensive information on a vast array of topics and concepts. This allows it to understand and engage with prompts across diverse domains. The user aims to steer the LLM through multiple rounds of interaction, activating relevant knowledge in a back-and-forth process. From this viewpoint, an adept LLM should function comparably to a sampler endeavoring to canvas the entire distribution of a particular concept or domain of knowledge that the user seeks to explore. In each interaction round, the LLM can be seen as drawing samples from a probability distribution defined over \(\mathcal{Z}\) which is conditioned on the prior prompts and responses. The goal of PE is for the LLM to facilitate a sequence of samples that approaches the true distribution of the user's targeted knowledge. This is an idealized assumption, considering the practical limitations, but it offers a clear objective for the development of LLMs in the context of PE: to refine their sampling process to better approximate the desired knowledge distribution through iterative prompting.

Regarding variability, it has been shown that LLM outputs depend heavily on the prompt provided. A superior LLM should produce varied, nuanced responses to small prompt variations. This allows prompt engineers to refine prompts iteratively, coaxing the LLM to generate high-quality, targeted responses. On the other hand, an adept LLM should indicate when a prompt lacks key information, rather than providing inconsistent outputs. This meta-cognitive capacity highlights areas for the user to enhance the prompt through subsequent iterations.

In summary, the immense knowledge and inherent variability of LLMs enable PE to unlock their potential through iterative refinement. However, these properties still lacks rigorous study. More investigation into these properties will undoubtedly lead to more efficient PE methods.

\subsubsection*{Challenges}

By examining the issues related to the three key tasks of PE listed in Section \ref{sec:sub:general_framework}, we gain a clearer understanding of the challenges for PE in both formulating problems and optimizing the prompting process. 

For problem formulation of PE, the complexities often arise from the discrete and structured nature of the language set \( \mathcal{Z} \). Both the evaluation function \( f \) and the prompt candidate sets \( \mathcal{P}_t \) are defined over \( \mathcal{Z} \). Given that \( \mathcal{Z} \) is a discrete space that possesses its own intricate linguistic structure, devising rule-based manipulations becomes inherently challenging. In addition, the changing of the prompt candidate set  \( \mathcal{P}_t \) with time $t$ adds another layer of complication in analyzing the underlying optimal control problem. Furthermore, it is often difficult to provide explicit definitions for \( f \) and \( \mathcal{P}_t \), which presents obstacles to the progress of PE in various fields.

With respect to optimizing the prompting process, the lack of access to the LLM's internal parameters necessitates the use of gradient-free optimization techniques such as random search or reinforcement learning. The efficiency of these methods is problematic, especially when factoring in the computational cost of interaction with the LLM. Additionally, the prompt candidate set \( \mathcal{P}_t \) often comprises an extensive array of potential prompts, thus inflating the action space for the optimal control problem considerably. This extensive action space further exacerbates the challenge of solving the optimal control problem, more so when coupled with the constraint of gradient-free optimization.

Despite these complexities, extensive empirical studies indicate that the quality of PE significantly influences the performance of LLMs \cite{liu2022generated, dai2023gpt, jung2022maieutic, chen2022program}. An intriguing question then arises: what are the limits to the effectiveness of PE? Amid these challenges and complexities lies a fertile ground for future investigation.

\section{Prompt Engineering Methods}
\label{sec:presentMethods}

In this section, we direct our attention towards some specific PE methods in the literature. As detailed in Section \ref{sec:sub:general_framework}, PE encompasses three pivotal elements: the evaluation function \( f \), the prompt candidate set \( \mathcal{P}_t \), and methods for solving the optimal control problem. Given that the evaluation function \( f \) is highly task-specific and subject to substantial variations, we opt not to concentrate on it within the scope of this section. Instead, our primary interest lies in the latter two elements: the mechanisms for enlarging \( \mathcal{P}_t \) and the algorithms capable of solving the ensuing optimal control problem.

Accordingly, in this section, we feature several notable methods pertinent to these two aspects. For each aspect, we first describe the task at hand and offer an interpretation of the highlighted PE methods within the context of our proposed optimal control framework. Following this, we present some insights that can be garnered by examining these PE methods through the lens of the proposed framework.

\subsection{Enlarging the Prompt Candidate Set}
\subsubsection{Enlarging via Previous Responses}

A branch of multi-round PE methods \cite{weng2023large,zheng2023progressive} enlarge their prompt candidate set by adopting previous responses as part of later prompts. Here, we use Progressive-Hint Prompting (PHP) \cite{zheng2023progressive} as an example to illustrate how these methods work.

PHP concentrates on arithmetic tasks. The evaluation function \(f\) is an identification function, signaling the correctness of the provided answer (e.g., having value 1 when the answer is correct and 0 otherwise).

\begin{figure}[!htbp]
\centering
\includegraphics[width=0.9\linewidth]{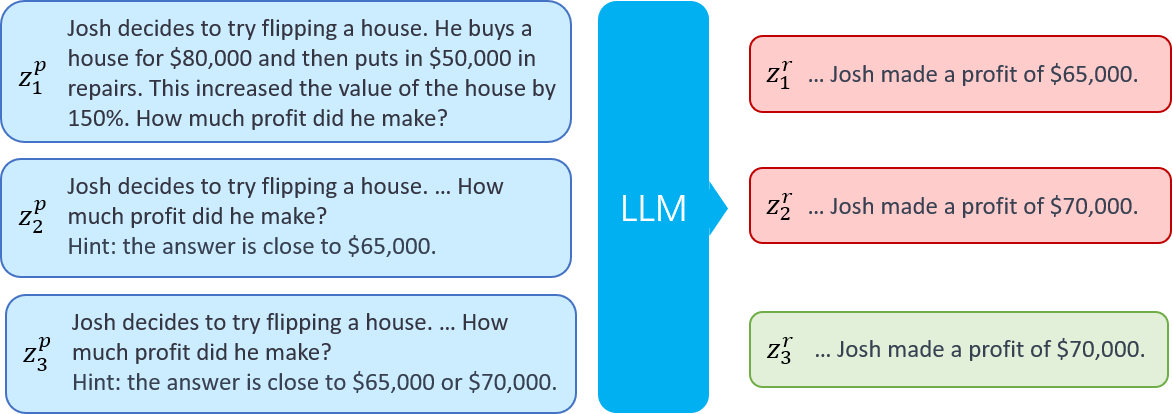}
\caption{
An example of PHP interactions. The subsequent prompt $z^p_{t+1}$ is constructed based on the previous round's prompt $z^p_t$ and the response $z^r_t$, by appending $z^r_t$ to the end of the hint array $z^p_t$.
Only when $z^r_1=$"\dots 730391." is present can $\mathcal{P}_2$ include prompts such as "\dots the answer is close to 730391." More generally, $\mathcal{P}_{t}=\mathcal{P}_{t-1}\cup\{z^p_t \text{ motivated by } z^r_1,\dots,z^r_{t-1}\}$.
}
\label{fig1}
\end{figure}

In PHP, the previous outputs of the LLM are used to construct subsequent inputs. Mathematically, this is given by
$$
z_t^{p} = [\text{\textbf{Question} ``Hint: the answer is close to } z^{r}_{1}, \dots, z^{r}_{t-1}"].
$$
The stopping criterion \(\tau\) is formulated as follows: terminate if \(z_\tau^{r}\) reiterates a portion of the prompt. This stopping rule relies on the heuristic notion that the correct answer, when present in the prompt, likely leads to an output that adheres to that answer. We give an example of PHP interactions in Figure $\ref{fig1}$. 

PHP-like PE methods review the responses from the LLM and utilize them to formulate subsequent prompts. These subsequent prompts may serve to clarify or provide additional guidance to the LLM, creating a feedback loop interaction paradigm that enhances adaptability across different LLMs and tasks.

\subsubsection{Enlarging via Direct Prompts}

Some multi-round PE methods utilize specially crafted prompts to expand their prompt candidate sets. These tactics may include directing the LLM to decompose the initial task \cite{zhou2023leasttomost, lei2023selfzcot}, or eliciting background information from the LLM \cite{Zhou2022LargeLM}, among other approaches. Although such interactions do not directly influence the final output, they effectively augment the prompt candidate sets, thereby aiding the generation of more impactful subsequent prompts. To demonstrate this approach, we refer to the Least-to-Most (LtM) method \cite{zhou2023leasttomost} as an illustrative example.

\begin{figure}[!htbp]
\centering
\includegraphics[width=0.9\linewidth]{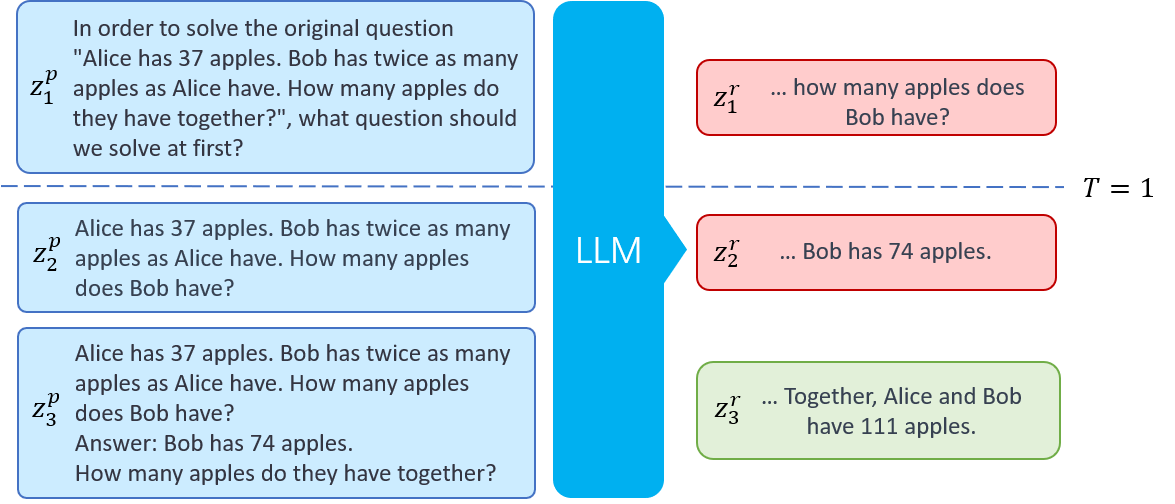}
\caption{
An example of LtM interactions with $T=1$. LtM employs the template "In order to solve \dots, we have to solve \dots" to break the original task into sub-tasks $\{z_t^r\}_{t=1}^{T}$. These sub-tasks contribute to the construction of $\mathcal{P}_{T+1}=\mathcal{P}_1\cup\{z^p \text{ related to }z^{r}_1,...,z^r_{T}\}$, which makes it a better prompt candidate set comparing with $\mathcal{P}_1$. For $t\geq T+1$, $\mathcal{P}_{t+1}$ consists of the intermediate results obtained during the dialog interactions starting from $t=T+1$. 
}
\label{fig2}
\end{figure}

LtM is primarily oriented towards reasoning tasks. It assume that the task $z^q$ takes the form of [\textbf{Description} \textbf{Question}]. Similar to PHP, LtM employs the identification function as its evaluation function \(f\). The LtM strategy employs the LLM to break down a complex task into simpler sub-tasks by using specially crafted initial prompts. Formally, LtM defines
\[
z^{p}_t = [\text{\textbf{Description}\ "In order to solve } z^{r}_{t-1}\text{, we have to solve:"}], \quad t = 1, \dots, T,
\]
with $z_{0}^{r}$ setting as the original question: \textbf{Question}. Additionally,
\[
z^{p}_t = \begin{cases}
    [\textbf{Description}\ z^{r}_{T}], & t=T+1 \\
    [z^{p}_{t-1}\ z^{r}_{t-1}\ z^{r}_{2T+1-t}], & t = T+2, \dots, 2T+1
\end{cases}.
\]
In the first \(T\) rounds, LtM uses the prompt "In order to solve \dots, we have to solve \dots" to iteratively dissect the original task \(z^q\) into sub-tasks \(\{z_t^{r}\}_{t=0}^{T}\). Then, from rounds \(T+1\) to \(2T+1\), it addresses these sub-tasks in a dialogue format and in reverse order. We give an example of LtM interactions in Figure $\ref{fig2}$. 

Creating a rule-based decomposition of the original task is challenging, which means it's difficult to formulate an initial prompt candidate set \(\mathcal{P}_1\) that encompass sub-tasks \(\{z_t^{r}\}_{t=0}^{T}\). Nevertheless, through the first \(T\) interactions and a fixed prompt template, LtM effectively expands its prompt candidate set \(\mathcal{P}_{T+1}\) to include all these sub-tasks, which makes $\mathcal{P}_{T+1}$ a much better prompt candidate set comparing with $\mathcal{P}_1$.

Both PHP and LtM use previous responses to generate future prompts, but they differ significantly in their strategies for expanding the prompt candidate set. The prompt "In order to solve \ldots, we have to solve \ldots." used by LtM may not directly contribute to completing the final task but facilitates a deeper understanding of the task, which in turn enhances the quality of subsequent prompts. While this strategy enriches the understanding of the task, it also introduces new complexity and potential misdirection if the decomposition does not align well with the final task’s requirements. Thus, the strategy offers both advantages and challenges.

\subsubsection{Insights and Future Directions}

Building upon the PE methods such as PHP and LtM, which employ an evolving \( \mathcal{P}_t \) for prompt candidate sets, the implications for optimal control theory are substantial. One compelling direction for future research lies in the theoretical formulation of non-stationary action spaces, which deviates notably from the assumptions of traditional optimal control frameworks. The conventional models often presuppose a static set of controls or actions, whereas the concept of an evolving \( \mathcal{P}_t \) introduces new layers of complexity and richness into the system dynamics. This demands a reevaluation of existing mathematical tools, from optimality conditions such as the Hamilton-Jacobi-Bellman equation to traditional concepts like stability and convergence. 

On the algorithmic front, a dynamically evolving \( \mathcal{P}_t \) presents intriguing challenges as well as opportunities. For instance, real-time adaptation becomes crucial; as an AI agent dynamically uncovers new actions or strategies, efficient algorithms are needed to incorporate these changes online. Additionally, a dynamic \( \mathcal{P}_t \) could exacerbate computational complexities, warranting new numerical methods that can adaptively optimize the system's behavior. This setting offers an interesting twist to existing methodologies like reinforcement learning, which typically operate in fixed action spaces, although some existing studies have already explored evolving action spaces \cite{farquhar2020growing, elman1993learning}. By allowing the action space to evolve, one could model more complex, adaptive behaviors. 

Therefore, the practice of dynamically updating \( \mathcal{P}_t \) as seen in PE methods in general not only adds empirical value to the task at hand but also poses sophisticated theoretical and algorithmic challenges. These challenges, in turn, offer fertile ground for extending the prevailing paradigms in optimal control theory to capture more intricate, evolving systems.

\subsection{Optimizing Prompts}

\subsubsection{Random Search}
Random search methods generate prompts at random and evaluate them using specific tasks to identify an optimal prompt \cite{Zhou2022LargeLM,yao2023tree,besta2023graph}. There is a wide variety of random search methods available. Here, we have chosen Automatic Prompt Engineering (APE) \cite{Zhou2022LargeLM} as an illustrative example.

APE aims to find an optimal prompt within all candidate prompts the LLM can provide. In other words, 
$$
\mathcal{P}_1=\{\text{all possible prompts LLM can generate based on } z^q\}.
$$ APE independently generated $N$ proposed prompts, forming a sub-candidate set $\hat{\mathcal{P}} \subset \mathcal{P}_1$ as follows:
\[
\hat{\mathcal{P}}=\{ z^{p,(k)} \overset{\mathrm{iid}}{\sim} \mathcal{D}_{\operatorname{LLM}},\ k=1,\dots, N \},
\]
where $\mathcal{D}_{\operatorname{LLM}}$ represents the prior distribution, which characterizes the distribution of proposed prompts generated by an independent LLM. APE evaluates these prompts $z^{p,(k)}$ through their scores $f^{(k)} = f(z^{p,(k)}; z^q)$ and discards the $M$ least effective prompts. Subsequently, $M$ new prompts are sampled to replenish $\hat{\mathcal{P}}$. The process is iterated until a termination criterion is met.

APE illustrates that a well-informed prior for $\mathcal{P}_t$, such as an LLM, can make random search a viable strategy for prompt optimization. However, the efficacy of this approach diminishes if the prior is less reliable, necessitating a large sample size and increased computational costs.

Additional examples include Tree of Thought (ToT) \cite{yao2023tree}, which specializes in multi-round PE. ToT utilizes a tree-search-like optimization strategy, systematically exploring sequences of prompts through iterative evaluation.

Random search methods are a promising tool for PE, and these methods could be improved to become even more efficient, effective, and versatile. For example, it is possible to develop new sampling algorithms that can quickly identify high-performing prompts, or to create hybrid methods that combine random search with other optimization techniques. Additionally, random search methods could be tailored to specific PE tasks, and better priors and evaluation metrics could be developed.

\subsubsection{Reinforcement Learning Methods}

In proposed optimal control framework for PE, it is crucial to recognize the inherent challenges in optimizing prompts for LLMs. A foremost challenge lies in the discrete nature of the underlying language space $\mathcal{Z}$, which does not lend itself to conventional optimization techniques that assume continuous spaces. Additionally, the dynamics governing the LLMs are often opaque; we lack access to the internal parameters and can treat these models only as black boxes. In such a complex setting, reinforcement learning has emerged as an especially effective approach to tackle these challenges.

Particularly, model-free RL algorithms align well with the black-box nature of LLMs. These algorithms, operating without knowledge of the underlying model, offer a viable strategy for optimizing control problems like those in PE. Methods like RLPrompt \cite{deng2022rlprompt} and PromptPG \cite{lu2023dynamic} are paradigmatic examples that adapt established RL techniques to optimize the quality of generated prompts. While these approaches show substantial promise, they necessitate numerous trial-and-error iterations, thereby elevating the computational overhead. This sets up an intriguing trade-off between the performance gains achieved and the computational resources expended, warranting a more thorough investigation.

Looking ahead, there is a multitude of directions for future research. One area deserving particular attention is the development of more sample-efficient RL algorithms that can achieve reliable performance without incurring prohibitive computational costs. Another avenue could involve devising hybrid methods that integrate domain knowledge into the RL framework, thereby potentially enhancing both the effectiveness and efficiency of the prompt optimization process.

\subsubsection{Comparing Random Search and Reinforcement Learning Methods}

The key distinction between random search methods and RL methods lies in the strategy for exploration and evaluation. RL methods, particularly model-free variants, operate on a principle of "evaluate and look ahead," allowing them to update their strategies based on the feedback received from prior interactions. This facilitates a more nuanced navigation of the prompt space, enabling RL to potentially find better prompts more efficiently. On the other hand, random search methods, such as APE and ToT, predominantly operate on a "generate and evaluate" paradigm without a look-ahead mechanism. They sample from a distribution, assess the samples, and make replacements, but do not typically leverage past evaluations to inform future explorations. While RL methods can incur higher computational costs due to their iterative nature and may require a well-defined reward function, random search methods are often simpler to implement and can be effective when a reliable prior is available. However, they may require a larger sample space and could be less efficient in navigating complex landscapes due to the lack of a lookahead mechanism. Thus, each approach comes with its inherent advantages and challenges, shaping their suitability for different PE scenarios.

\subsection{Discussions}

Existing PE methods have made valuable progress on certain challenges mentioned in Section 2.4, displaying ingenuity in solving key facets like optimization and prompt candidate set expansion. However, their focused approaches also reveal opportunities for further advancement. For instance, APE and ToT concentrate on a rather specific optimization approach lacking generalized strategies, while LtM and PHP expand prompt candidate sets for specific tasks without wider applicability. 

Within our framework \eqref{def:PE_basic}, optimization and prompt candidate set expansion could potentially be unified, enabling joint optimization instead of the isolated treatment seen in current methods. Additionally, task-specific strategies developed for existing techniques could be translated into more generalizable principles via the formal optimal control perspective offered by our framework.

By enabling a holistic, mathematically grounded understanding, our framework provides tools to elevate PE solutions to the next level of sophistication. We see great potential in moving from independent methods toward comprehensive techniques with broader applicability. We hope by adopting this systematic view, we can unlock the full capabilities of human-LLM interaction.

\section{Further Extensions}
\label{sec:extensions}

The optimal control framework given by \eqref{def:PE_basic} serves as a foundational structure for the mathematical description of numerous prevalent PE methods. Nevertheless, it is imperative to acknowledge that \eqref{def:PE_basic} is not universally adequate for capturing all aspects of the existing PE techniques and applications. In this section, we intend to delve into specific instances of certain advanced PE methodologies and articulate how they can be recast as optimal control problems.

\subsection{Prompt Engineering via Ensemble Methods}

In statistics and machine learning, ensemble methods have long been instrumental for augmenting predictive accuracy and robustness. These techniques leverage multiple instances of similar procedures to yield superior performance compared to single trials \cite{Opitz_1999,polikar2006ensemble,rokach2010ensemble}. Building on this established groundwork, ensemble techniques have been naturally adapted to the realm of PE, where they have yielded noteworthy outcomes \cite{Wang2022SelfConsistencyIC,fu2023complexitybased,li2023making,Sorensen_2022,Imani2023MathPrompterMR}.

By examining the optimal control framework \eqref{def:PE_basic}, it becomes evident that specialized PE methods like Self-Consistency CoT \cite{Wang2022SelfConsistencyIC} and Mutual Information \cite{Sorensen_2022} do not align seamlessly with the existing formalism. To address this, this subsection presents an adapted framework based on \eqref{def:PE_basic} to better accommodate these ensemble PE methods.

Consider a general human-LLM multi-query scenario for one task $z^q$. Each query is denoted by a prompt $z_i^p$ and the corresponding response $z_i^r$, where $i\in I$ is the query's index with $I$ being the index set. The final response $z^r$ to the task is formulated using an ensemble function $\operatorname{En}(\cdot)$ applied to all these responses:
\[
z^r=\operatorname{En}(\{z_i^{r}\}_{i\in I}),
\]
where $\operatorname{En}(\cdot)$ represents the ensemble strategy in use. Using $\omega_i$ to denote the randomness within the LLM, the optimal control problem for PE via ensemble methods can be formulated as:
\begin{equation}
    \begin{aligned}
        \max_{z_i^{p}\in\mathcal{P}} &  \, \mathbb{E}_{\omega_i}f(\operatorname{En}(\{z_i^{r}\});z^q)\\
        s.t.\  & z_i^{r}=\operatorname{LLM}(z_i^{p},\omega_i),\ i\in I\\
    \end{aligned}
    \label{def: PE_Ensembling}
\end{equation}
The mathematical formulation presented is sufficiently versatile to subsume a diverse array of ensemble PE methods.

In one class of approaches, a same prompt is utilized across multiple queries, leveraging the inherent stochasticity of the LLM to introduce variation. The specific ensemble function, denoted as \( \operatorname{En}(\cdot) \), further dictates the characteristics of the ensemble PE method. For instance, when employing \( \operatorname{En}(\cdot) \) as a majority-voting scheme, the method of Self-Consistency CoT \cite{Wang2022SelfConsistencyIC} is naturally encapsulated. Conversely, Complexity-CoT \cite{fu2023complexitybased} arises when \( \operatorname{En}(\cdot) \) is implemented as a complexity threshold.

An alternative avenue for ensemble construction involves the introduction of nuanced variations in the prompts across different trials. For instance, Mutual Information \cite{Sorensen_2022} uses disparate prompt templates to generate different $\{z^p_i\}$. MathPrompter \cite{Imani2023MathPrompterMR}, on the other hand, deploys two different classes of prompts (algebraic and Python prompts) to prompt the LLM. Further extending this notion, Step-Aware Verifier \cite{li2023making} recommends querying a single LLM with \( M_1 \) distinct (types of) prompts, each replicated \( M_2 \) times, thereby offering a generalized methodology that could be viewed as an extension of MathPrompter's approach. Empirical validation corroborates the effectiveness of these ensemble PE methodologies.

The inherent stochasticity of LLMs is sometime perceived as a drawback, particularly in applications where deterministic outputs are traditionally sought. However, ensemble PE methods compellingly illustrate that this stochastic nature can be exploited to advantageous ends. By introducing ensemble methods into the optimal control framework for PE, we can harness this stochasticity to improve performance, rather than treating it as an impediment.

One direction for future work is to extend the framework to include the function \( \operatorname{En}(\cdot) \) as an explicit control variable. By doing so, the framework could offer a systematic way to optimize ensemble strategies for specific LLMs and tasks. Whether the optimal strategy employs majority voting, complexity thresholds, or more nuanced mechanisms could be determined within this formalism, providing a unified metric for evaluation. Additionally, the optimal control framework supports adaptive selection of the ensemble strategy based on observed performance along with each additional query, allowing real-time fine-tuning of ensemble methods. 

\subsection{Prompt Engineering via Multi-Agent Collaboration}
In the context of LLMs, multi-agent systems refer to a collection of interactive agents that work collaboratively to achieve a collective objective.  An "agent" in this context is defined as an LLM operating under a given initial instruction. Different initial instructions yield distinct agents, which can display considerable heterogeneity in their behavior and capabilities. Each agent is responsible for generating prompts to facilitate interactions with each other. 

We extend our terminology to accommodate the intricacies of multi-agent PE. Denote $i\in I$ as the index for LLM-based agents. Then the prompt candidate sets are designated as \( \mathcal{P}_{(i, t)} \), the prompt for the \(i\)-th agent at time-step \(t\) is denoted as \( z_{(i, t)}^{p} \), and the corresponding response is \( z_{(i, t)}^{r} \). Here we extend the optimization target $f$ to $f_i$ as well to evaluate the PE quality for each corresponding agent.

The extended framework for the multi-agent PE is given as follows
\begin{equation}
\begin{aligned}
    \max_{\{\tau_i\}} &\ \ \max_{\left\{z_{(i, t)}^{p}\in \mathcal{P}_{(i, t)}\right\}}\ \ \sum_{i \in I}
    f_i(z_{(i, \tau_i)}^{r}; z_i^q) \\
    \text{s.t.} & \quad z_{(i, t)}^{r}=\operatorname{LLM}(z_{(i, t)}^{p}),\ i\in I.
\end{aligned}
\label{def:PE_multi-agent}
\end{equation}

The framework articulated by \(\eqref{def:PE_multi-agent}\) can be used to describe a variety of multi-agent PE techniques, among which a predominant focus is the automated problem solvers, exemplified by task-specific applications like automated program development \cite{qian2023communicative}, or general problem solving with improved factuality and reasoning \cite{du2023improving}. In these systems, each agent can be meticulously designed with a well-defined role, articulated through their initial prompts. This enables a collaborative environment in which agents, each specializing in a particular aspect of the problem at hand, work collaboratively to achieve an efficient and effective task completion. In the realm of social simulations, multi-agent PE serves to model complex interactions like trading negotiations or role-playing scenarios \cite{fu2023improving, junprung2023exploring}. Here, utility functions \( f_i \) are often tailored to assess the meaningfulness of the simulated dynamics, rather than achieving a specific task.

Embedding multi-agent PE within the framework of optimal control offers several advantages and insights. For instance, by conceptualizing each agent as an individual control unit guided by its utility function, one may gain a structured view for analyzing the coordinated actions and objectives of a multi-agent system. This allows for a mathematical description of how different agents, each with their unique initial prompt defining its role, contribute to the global objective, thereby unifying disparate approaches under a single mathematical umbrella.

\section{Conclusions}

In this paper, we have introduced an optimal control framework to describe multi-round prompt engineering. The framework is shown to be rather flexible, accommodating an expansive array of problem settings and objectives that frequently appear in the literature. The proposed framework has the potential to facilitate the development of more efficient PE algorithms, enabling more effective control over the LLMs and further broadening the scope of achievable tasks. Furthermore, the framework can be extended to incorporate ensemble methods and multi-agent scenarios. As discussed in various sections of this paper, the proposed frameworks grant a unified perspective on PE methods, has enabled us to propose various possible improvements to existing PE methods and has illuminated new directions for future research.

\section*{Acknowledge}
ZZ is supported by the National Key R\&D Program of China, Project Number 2021YFA1001200, and the NSFC, grant Number 12031013, 12171013. BD is supported by the NSFC, grant Number 12090022.

\bibliography{ref}

\end{document}